\pdfoutput=1

\documentclass[11pt]{article}

\usepackage[preprint]{acl}

\usepackage{times}
\usepackage{latexsym}
\usepackage{hyperref}

\usepackage[T1]{fontenc}

\usepackage[utf8]{inputenc}

\usepackage{microtype}

\usepackage{inconsolata}

\usepackage{graphicx}
\usepackage{booktabs}   
\usepackage{adjustbox}
\usepackage{multirow}
\usepackage{xcolor}         
\usepackage{tcolorbox}
\newcommand{\VarSty}[1]{\textnormal{\ttfamily\color{blue!90!black}#1}\unskip}
 
\title{IDA-VLM: Towards Movie Understanding via ID-Aware Large Vision-Language Model}

\author{
 \textbf{Yatai Ji\textsuperscript{1,2}},
 \textbf{Shilong Zhang\textsuperscript{1}},
 \textbf{Jie Wu\textsuperscript{3}},
 \textbf{Peize Sun\textsuperscript{1}},
\\
 \textbf{Weifeng Chen\textsuperscript{3}},
 \textbf{Xuefeng Xiao\textsuperscript{3}},
 \textbf{Sidi Yang\textsuperscript{2}},
 \textbf{Yujiu Yang \textsuperscript{2}}\thanks{\ \ Corresponding author.},
 \textbf{Ping Luo\textsuperscript{1}\footnotemark[1]}
\\
 \textsuperscript{1}The University of Hong Kong,\quad
 \textsuperscript{2}Tsinghua University,\quad
 \textsuperscript{3}ByteDance
}

\begin{document}
\maketitle
\begin{abstract}
The rapid advancement of Large Vision-Language models (LVLMs) has demonstrated a spectrum of emergent capabilities. Nevertheless, current models only focus on the visual content of a single scenario, while their ability to associate instances across different scenes has not yet been explored, which is essential for understanding complex visual content, such as movies with multiple characters and intricate plots. Towards movie understanding, a critical initial step for LVLMs is to unleash the potential of character identities memory and recognition across multiple visual scenarios. To achieve the goal, we propose visual instruction tuning with ID reference and develop an \textbf{ID}-\textbf{A}ware Large \textbf{V}ision-\textbf{L}anguage \textbf{M}odel, IDA-VLM. Furthermore, our research introduces a novel benchmark MM-ID, to examine LVLMs on instance IDs memory and recognition across four dimensions: matching, location, question-answering, and captioning. Our findings highlight the limitations of existing LVLMs in recognizing and associating instance identities with ID reference. This paper paves the way for future artificial intelligence systems to possess multi-identity visual inputs, thereby facilitating the comprehension of complex visual narratives like movies. 
Codes and models are available at \url{https://github.com/jiyt17/IDA-VLM}.
\end{abstract}

\section{Introduction}

Our real world contains a wide variety of information, such as texts, images, sounds, etc. 
Towards multimodal comprehension~\cite{DBLP:journals/access/GuoWW19/survey, DBLP:conf/iclr/LuCZMK23/unifiedio}, inspired by the success of Large Language Models (LLMs)~\cite{chatgpt,vicuna,DBLP:journals/corr/abs-2302-13971/llama,llama3,bard}, there is a surging interest in Large Vision-Language Models (LVLMs), such as LLaVA~\cite{DBLP:conf/nips/LiuLWL23a/llava}, Otter~\cite{DBLP:journals/corr/abs-2305-03726/otter}, GPT-4V~\cite{gpt4v}, Gemini~\cite{gemini}, and others. In the quest for Artificial General Intelligence (AGI), LVLMs serve as pivotal milestones, enhancing machines' capabilities in multimodal perception, reasoning, and knowledge. 

\begin{figure}[t]
	\centering
	\includegraphics[width=0.5\textwidth]{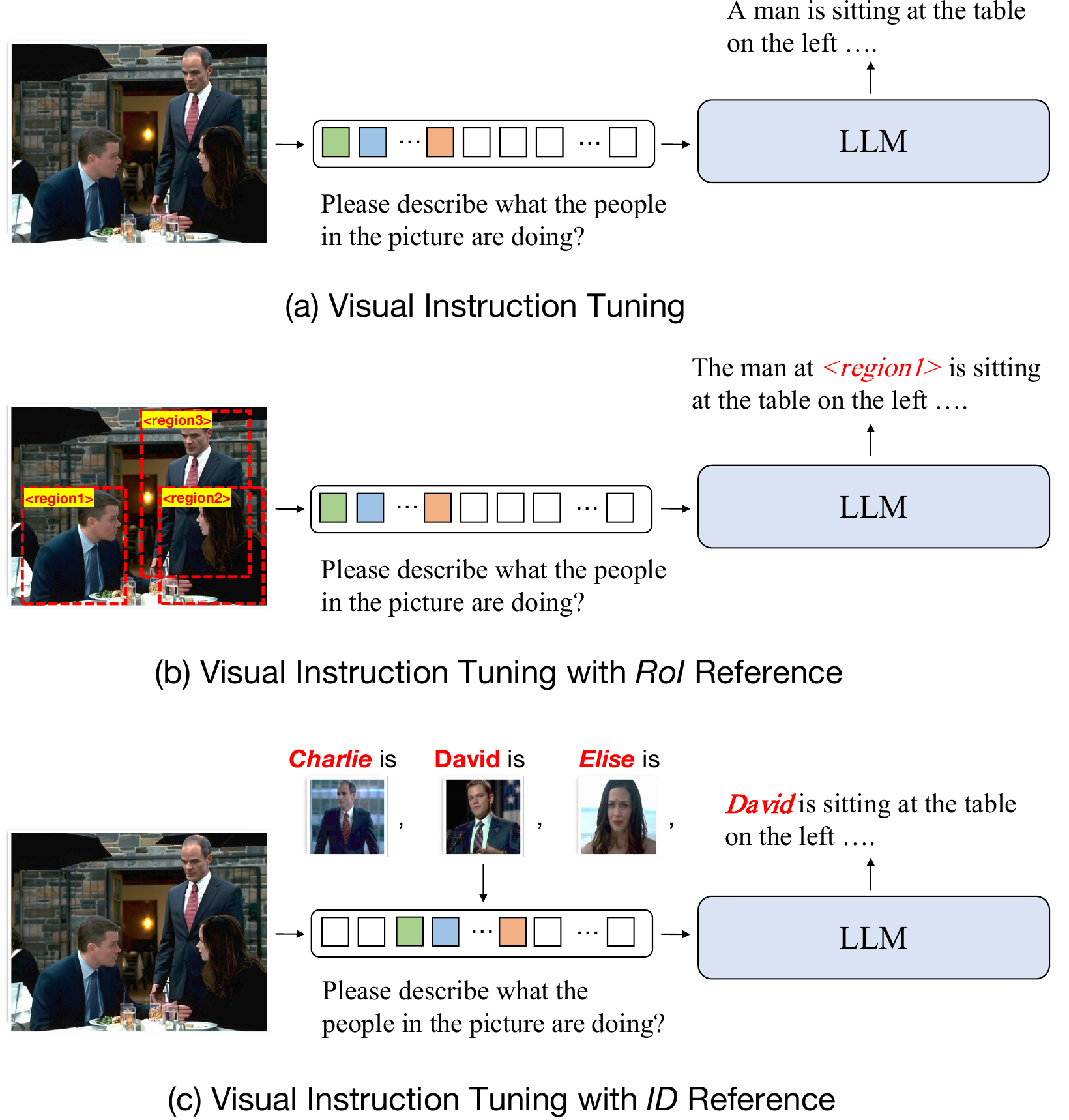}
 
 \caption{
 \textbf{Comparison of different Visual Instruction Tuning Formats.} 
 For visual instruction tuning with ID reference, we arrange the names and images of each character as references. The model should be able to recognize the correct character identity and then answer the user's instructions.}
	\label{fig:intro compare}
 \vspace{-0.3cm}
\end{figure}

Typical LVLMs incorporate visual encoders with LLMs, such as LLaVA~\cite{DBLP:conf/nips/LiuLWL23a/llava}, MiniGPT4~\cite{DBLP:journals/corr/abs-2304-10592/minigpt4}. These models project an image's visual features into the embedding space of language models and employ visual instruction tuning, allowing users to complete a variety of visual tasks through language instructions, as depicted in Figure \ref{fig:intro compare}(a). However, pure language interaction makes VLMs hard to receive precise Region-of-Interest references from users, thus hindering their capability to focus on specific regions of an image. To address this issue, a series of works~\cite{DBLP:journals/corr/abs-2306-15195/shikra, DBLP:journals/corr/abs-2307-03601/gpt4roi, DBLP:journals/corr/abs-2312-00784/vipllava, DBLP:journals/corr/abs-2310-09478/minigptv2, DBLP:journals/corr/abs-2307-09474/chatspot, DBLP:journals/corr/abs-2311-03356/glamm} attempts to allow adding RoI references to language instructions and then use region-level data to align these RoI references to the image. As shown in Figure \ref{fig:intro compare}(b), these excellent works usually support using position coordinates or visual prompts as the RoI reference, which provides a more flexible interactive experience within specific areas.

Previous LVLMs have demonstrated versatile capabilities for visual understanding from the image level to Region-of-Interest. However, these models can only process the visual input of a single scenario, and their ability to associate instances across different scenes has not yet been explored, which requires the model to memorize the instance identity and recognize it in different scenes. This ability, which we call `ID awareness', is a fundamental human visual competency essential for comprehending complex multi-identity visual inputs, such as movies and animations. In the context of movie understanding, viewers should remember the name and appearance of each character, then recognize them across disparate scenes correctly, linking instances across diverse images for accurate narrative interpretation.

To achieve ID awareness in LVLMs like humans, as shown in Figure \ref{fig:intro compare}(c), we propose visual instruction tuning with ID reference, which allows users to utilize the name and a reference image of the character (ID image) to define the identity in the prompt, and raise questions about test images.  
However, there is no off-the-shelf visual tuning data with ID reference, so we meticulously craft a dual-stage instruction tuning dataset based on existing datasets. 
The initial phase leverages annotations in datasets such as VCR~\cite{DBLP:conf/cvpr/ZellersBFC19/vcr}, Flickr30k~\cite{DBLP:journals/ijcv/PlummerWCCHL17/f30k}, and RefCOCO~\cite{DBLP:conf/emnlp/KazemzadehOMB14/refcoco} with our data configuration strategies, to tutor the model on associating the instances of ID images with test images, where ID references are extracted from the test image. The subsequent phase utilizes MovieNet~\cite{DBLP:conf/eccv/HuangXRWL20/movienet} 
to generate Q\&A and caption instruction tuning data with GPT-4V~\cite{gpt4v}. The second stage data contains ID references of higher quality, which is more similar to real movie understanding. 
These tuning data unleash the potential of LVLMs to memorize and recognize instance identities in different scenes. 
Based on the visual instruction tuning with ID reference, we develop an ID-aware LVLM, called IDA-VLM, suitable for multi-identity visual comprehension. 
Our model learns the fine-grained identity information from ID references and generalizes to recognize characters from different scenes of test images. 
We also introduce a specialized component, termed ID-Former, to enhance the model's capability of recognizing character identities. 

To measure ID recognition capability for movie understanding, we propose a new benchmark, namely MM-ID. 
Our MM-ID aims at challenging LVLMs to remember and recognize instance identities, further understanding complex visual scenes. MM-ID dissects LVLMs' competency across four progressively complex levels: matching, location, question-answering, and captioning. 
MM-ID comprises a collection of 585 diverse testing samples, whose instance identities source from the characters of movies, the roles of animations, and individualized animals and objects. We test both open-source models and closed-source APIs, which exposes their apparent deficiencies in identity memory and recognition, and the results are unsatisfactory even for GPT-4V. The evaluation results not only reveal the current limitations of LVLMs in instance recognition with ID reference but also highlight our benchmark’s significance. In contrast, IDA-VLM achieves the best performance on MM-ID, which demonstrates the effectiveness of our instruction tuning strategies and model design.

In a nutshell, our contributions are summarized as follows:
\begin{itemize}
\item This paper is the first attempt to investigate ID awareness of LVLMs for complex multi-identity visual comprehension, like movies. We propose visual instruction tuning with ID reference and construct corresponding tuning datasets. 
\item We develop an ID-aware LVLM, IDA-VLM to recognize character identities and adopt a dual-stage fine-tuning method to unleash the potential of LVLMs in identity memory and recognition across diverse scenes.
\item We propose MM-ID, a novel benchmark to examine LVLMs on identity memory and recognition. We evaluate representative LVLMs on our benchmark, uncovering the limitations of existing LVLMs in contextual identity recognition. Furthermore, our model achieves State-of-The-Art (SoTA) performance on MM-ID.
\end{itemize}

\section{Related Work}

\subsection{Large Vision-Language Models}

Conventional multimodal models consist of uni-modal encoders and cross-modal fusion encoders. Relying on vision-language pre-training~\cite{DBLP:conf/emnlp/TanB19/lxmert, DBLP:conf/nips/LuBPL19/vilbert, Dou_2022_CVPR/meter, DBLP:conf/emnlp/WangJSYS21/mirtt, DBLP:conf/cvpr/JiWGZZWZSY23/map}, they have shown an impressive cross-modal semantic alignment ability~\cite{DBLP:conf/cvpr/JiTJKCZWY023/scl, DBLP:journals/corr/abs-2306-07096/glscl}, which brings substantial advances on various downstream tasks~\cite{balanced_vqa_v2/vqa_v2, DBLP:journals/ijcv/PlummerWCCHL17/f30k, DBLP:conf/eccv/LinMBHPRDZ14/mscoco}. However, due to the limitations of model size and training data scale, the performance of vision-language pre-training models is unsatisfied in open-ended scenarios.

Nowadays, Large language model (LLM) has exhibited remarkable abilities to understand, reason, and generate texts. Large Vision-Language Model (LVLM)~\cite{DBLP:journals/corr/abs-2311-12793/sharegpt4v, DBLP:conf/nips/Dai0LTZW0FH23/instructblip, DBLP:journals/corr/abs-2401-16420/intern} incorporates visual encoder and LLM, aligning visual features to the embedding space of LLM. Leveraging strong generalization and emergent capability of LLM, LVLM realizes free multimodal interactions with human. 
Flamingo~\cite{DBLP:conf/nips/AlayracDLMBHLMM22/flamingo} is a pioneering work on extending LLMs to vision-language pretraining by inserting additional cross-attention layers for visual input. 
BLIP-2~\cite{DBLP:conf/icml/0008LSH23/blip2} proposes Q-former to map the visual features to the hidden space of language models. 
To date, various works have shown encouraging progress with instruction tuning, including MiniGPT-4~\cite{DBLP:journals/corr/abs-2304-10592/minigpt4}, LLaVA~\cite{DBLP:conf/nips/LiuLWL23a/llava}, Otter~\cite{DBLP:journals/corr/abs-2305-03726/otter}, which demonstrate impressive results on natural instruction-following and visual reasoning capabilities. 
The perceptual capabilities of LVLMs are evolving towards fine-grained understanding. 
Numerous models~\cite{DBLP:journals/corr/abs-2307-03601/gpt4roi, DBLP:conf/nips/LiuLWL23a/llava, DBLP:journals/corr/abs-2306-15195/shikra, DBLP:conf/nips/WangCCWZZLLZQD23/visionllm} focus on region-level understanding, using visual instruction tuning with RoI reference, so that users could ask questions about specific instances within the content. 
Moreover, there are some LVLMs equiped with Stable Diffusion, which can produce multimodal outputs~\cite{DBLP:conf/nips/KohFS23/gill, DBLP:journals/corr/abs-2310-01218/seed,DBLP:journals/corr/abs-2307-05222/emu,DBLP:journals/corr/abs-2309-08637/textbind}. 
LVLMs hold the potential to perform a wider array of functions and to understand more intricate visual information. 
For instance, movie is considered as one of the most intricate mediums for conveying visual information. 
This paper researches on movie understanding with ID reference, recognizing characters across disparate scenes.



\subsection{Multimodal Benchmark}
Previous evaluation for multimodal models measures some specific abilities, such as VQAv2~\cite{balanced_vqa_v2/vqa_v2}, GQA~\cite{DBLP:conf/cvpr/HudsonM19/gqa} for visual question answering, RefCOCO~\cite{DBLP:conf/emnlp/KazemzadehOMB14/refcoco} for visual grounding, Visual7W~\cite{DBLP:conf/cvpr/ZhuGBF16/v7w} for PointQA, VCR~\cite{DBLP:conf/cvpr/ZellersBFC19/vcr} for commonsense reasoning in an image. 
Recently, there is a surging interest in developing comprehensive benchmarks for evaluating LVLMs. 
MMBench~\cite{DBLP:journals/corr/abs-2307-06281/mmbench} consists of multiple-choice questions and introduces a CircularEval strategy for evaluation framework. 
LLaVA-Bench~\cite{DBLP:conf/nips/LiuLWL23a/llava} employs GPT-4 to assess responses from both GPT-4 and the model under evaluation, then provides scores and explanations for the answers. 
MM-Vet~\cite{DBLP:journals/corr/abs-2308-02490/mmvet} assesses LVLMs across six fundamental visual-linguistic capabilities, utilizing a GPT-4-based evaluator for open-ended responses. 
MME~\cite{DBLP:journals/corr/abs-2306-13394/mme} tests for perception and cognition competencies across 14 distinct subtasks, which requires models to provide simple `yes' or `no' answers. 
SEED-Bench~\cite{DBLP:journals/corr/abs-2307-16125/seedbench} covers 12 evaluation dimensions and adopts an answer ranking strategy to evaluate LVLMs via multiple-choice questions. 
POPE~\cite{DBLP:conf/emnlp/LiDZWZW23/pope} is a dedicated benchmark for assessing object hallucination. 
Existing benchmarks evaluate multifaceted performance of LVLMs, yet they predominantly focus on visual scenarios that lack character identities and intricate plots. 
 
\section{Method}

\begin{figure*}[t]
	\centering
	\includegraphics[width=\textwidth]{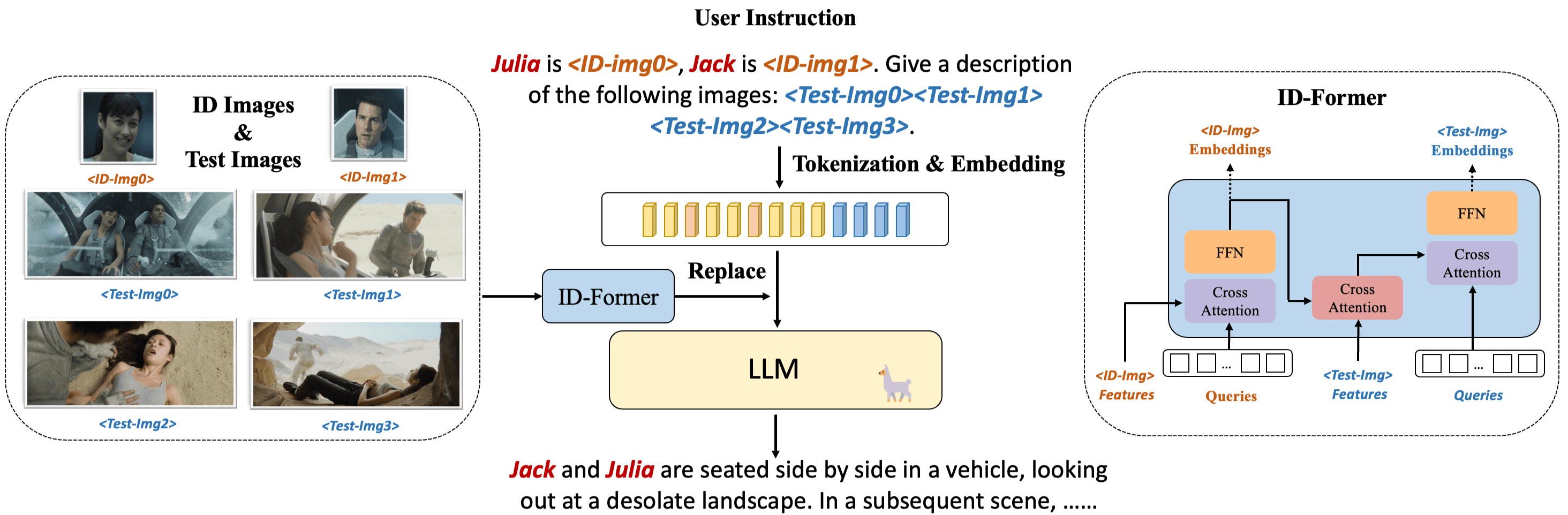}
	\caption{IDA-VLM is an end-to-end vision-language model for processing instructions that contain references to specific instances. We introduce ID reference as using a correlating character reference image and corresponding name to characterize an identity, exemplified as \textit{Julia is <ID-Img\{i\}>}. During tokenization and conversion to embeddings, the embedding of \textit{<ID-Img\{i\}>} and \textit{<Test-Img\{i\}>} in the instruction are replaced with the ID and Test image embeddings respectively. A simple yet effective image feature projector termed ID-Former is proposed to enhance the ID identification ability. As the output in the figure, IDA-VLM can memorize these character IDs, recognize them in test images, and respond to user instructions with the correct ID references. }\label{fig:model achitecture}
 \label{fig:architecture}
 \vspace{-0.3cm}
\end{figure*}

To understand complex visual input, for example, a movie or animation, the model need memorize and recognize character identities, linking characters in distinct scenes. 
As shown in Figure \ref{fig:architecture}, we simplify movie understanding as given ID images of certain characters and keyframes from the movie or animation served as test images, the model completes instruction tasks. The ID images, test images, and instruction text are sent to the model together for inference in an interleaved format of images and text. 
The model need to memorize instance identities in ID references and recognize them in test images for providing responses. 

\subsection{Model Architecture}

In this paper, we adopt Qwen-VL-chat~\cite{DBLP:journals/corr/abs-2308-12966/qwen} as our baseline model. 
In order to adapt the model to the task of instance ID recognition, we employ a dual-stage visual instruction tuning with ID reference. 
The architecture of IDA-VLM comprises three components: a visual encoder, ID-Former consisting of cross-attention mechanisms, and a subsequent large language model. 

As shown in Figure \ref{fig:architecture}, the ID-Former is designed to project visual features into the input semantic space of LLM and contribute to recognizing instance identities. This is achieved by two cross-attention modules. The first one interacts learnable queries with the visual features through cross-attention, effectively compressing the visual semantics into a shorter, fixed-length feature encoding. The second cross-attention utilizes queries of ID images to modulate test image embeddings, activating identity information of test images.

\subsection{First-stage Tuning}

In the first phase, we harness the off-the-shelf annotations available in existing datasets along with our proposed data configuration strategies, reducing the cost for extra annotations about ID recognition. Specifically, we utilize the public datasets containing instance spatial information, including Visual Commonsense Reasoning (VCR)~\cite{DBLP:conf/cvpr/ZellersBFC19/vcr}, RefCOCO~\cite{DBLP:conf/emnlp/KazemzadehOMB14/refcoco}, and Flickr30k~\cite{DBLP:journals/ijcv/PlummerWCCHL17/f30k} datasets, to construct visual instruction tuning data with ID reference, which contains three instruction tasks: question-answering (Q\&A), location, captioning.

\begin{figure*}[t]
	\centering
	\includegraphics[width=\textwidth]{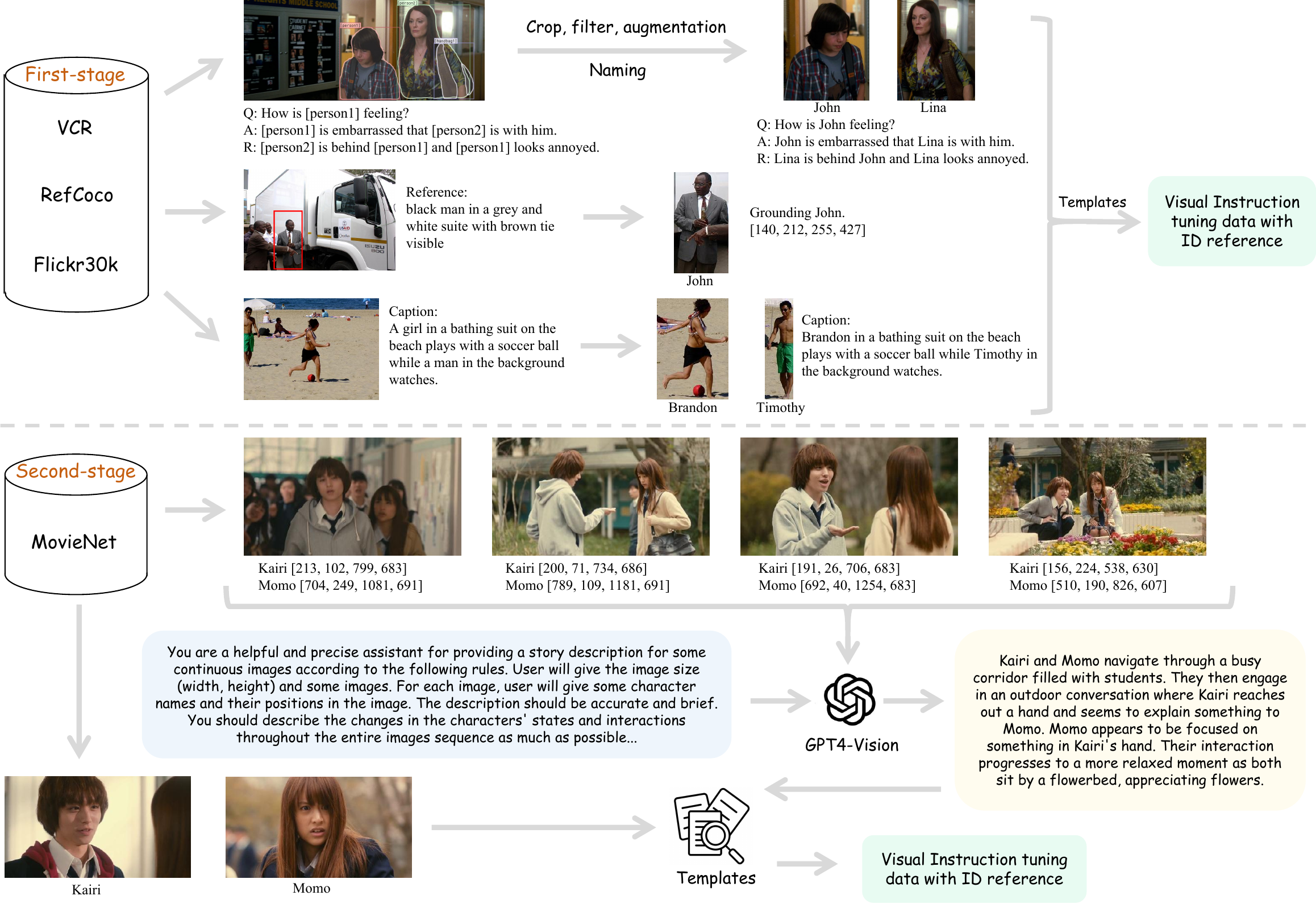}
	\caption{First-stage and second-stage instruction tuning data construction pipelines.}
	\label{fig:anno pipeline}
 \vspace{-0.3cm}
\end{figure*}

VCR dataset consists of questions that require commonsense reasoning to answer, as well as choosing the reasons. These questions ask about particular people or objects, which have location annotations. We crop sub-images of the individuals and assign them identifiable labels, such as person names. By replacing the people in the questions and answers with these labels, we gain a new set of QA data that necessitates ID recognition. 
RefCOCO (and variants RefCOCO+ and RefCOCOg) is a dataset targeted on visual localization based on descriptions. For the items to be localized, we created ID images after cropping them from the original images to perform localization tasks. Flickr30k, a dataset comprising image descriptions, includes coordinates for the people or objects mentioned in the descriptions. We extracted the areas associated with individuals to produce ID images and provided names to generate corresponding descriptions. 
Additionally, we conduct some data augmentation and cleaning operations, such as filtering out samples with only one person or those with sub-images of inappropriate sizes.
The annotation pipeline is shown in the upper part of Figure \ref{fig:anno pipeline}.

The essential purpose of this stage is to train the model on its ability to build relationships between multiple images, using the ID images to identify and accurately locate or describe objects within test images. 
However, the instance ID images are cropped from the original images, which reduces the difficulty of recognition. 

\subsection{Second-stage Tuning}

The second-stage fine-tuning data is based on the MovieNet~\cite{DBLP:conf/eccv/HuangXRWL20/movienet} dataset. The original annotations in the MovieNet dataset encompass the names of the characters present in each movie shot and their coordinate location information. In the MovieNet, we select pictures that contain only a single character to serve as ID images, while those containing multiple characters are used as test images. This approach allows us to naturally develop a dataset for location task. Moreover, we compile ID images featuring the same character to construct data for matching task.

For Q\&A and caption tasks, we adopt GPT-4V to convert annotations from the MovieNet dataset into the question format that refers to specific roles. Specifically, as illustrated in the lower part of Figure \ref{fig:anno pipeline}, we feed test images along with their character location information into GPT-4V, and encourage the model to generate captions or question-answer pairs via prompt engineering. The detailed prompts used for generating descriptions or question-answer pairs are shown in Appendix \ref{tuning data construction}. Finally, we integrate ID images, test images and results of GPT-4V into conversation templates, producing second-stage instruction tuning data. 

After completing data construction process, we train our model with next-token prediction loss in two stages. 
The first stage tuning data contains approximately 200,000 samples, while the second one comprises around 60,000 samples. 
Moreover, to ensure that the model would not forget its previously learned knowledge during ID recognition training, we integrate instruction tuning data from LLaVA with the aforementioned data, training them simultaneously.

\section{MM-ID}

\subsection{Problem Definition}

\begin{figure*}[t]
	\centering
	\includegraphics[width=\textwidth]{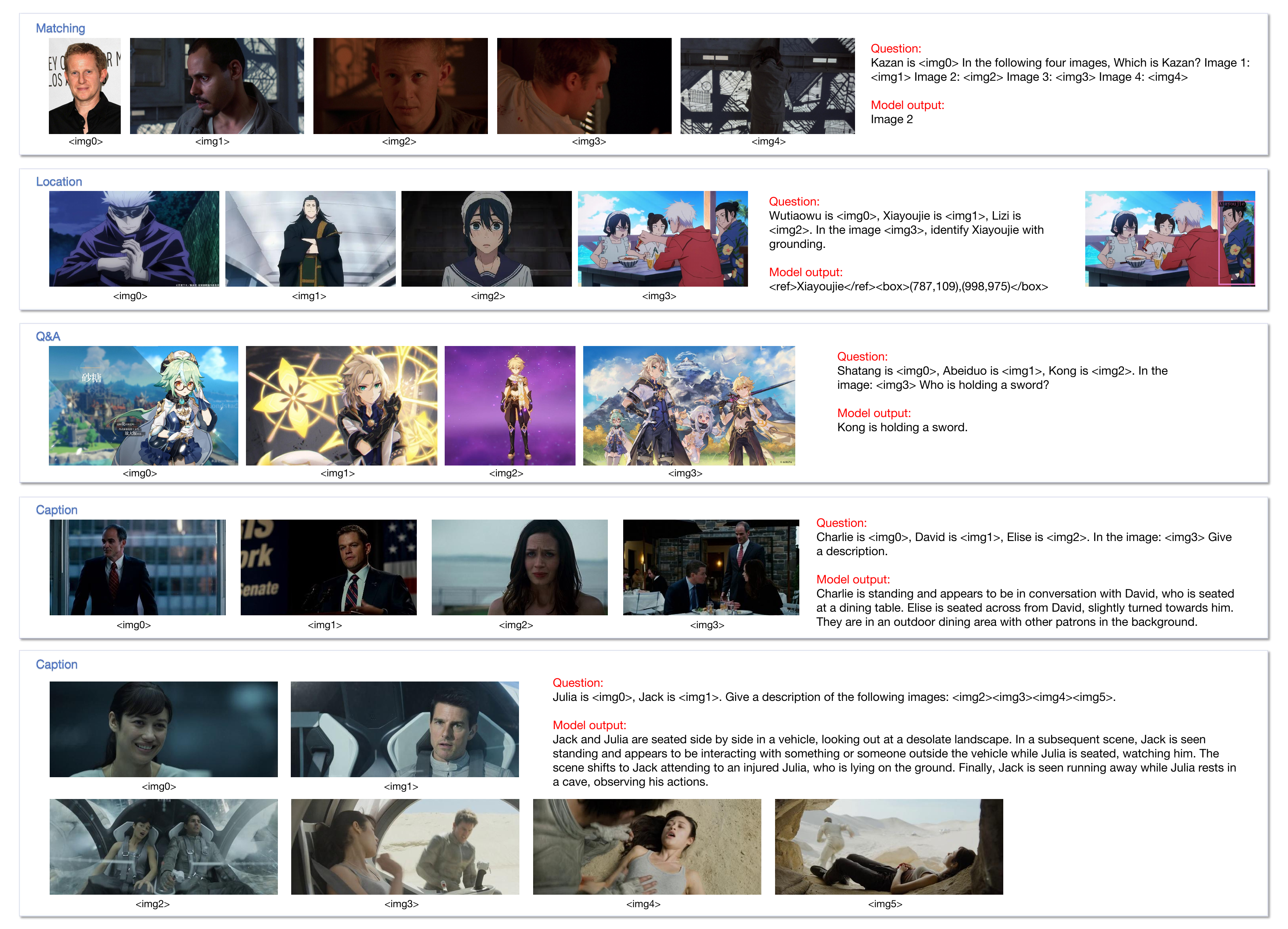}
	\caption{Data samples of MM-ID\protect\footnotemark[1]. Each sample consists of ID images of each character, and test images containing multiple characters.  Our model can memorize the identity information in ID reference and generalizes to recognize characters from different scenes.}
	\label{fig:benchmark}
 \vspace{-0.2cm}
\end{figure*}

To measure ID recognition capability of models, we propose a new benchmark, MM-ID. We appraise the capability of models to recognize IDs across four incrementally complex levels. The first sub-task we investigate is matching: given an instance image, which could feature a person, an animal, or a building, the model must choose an image from four options that contains the same instance. 
The second sub-task involves localizing the instance within a test image based on the ID image, and providing coordinates of the bounding box, with the challenge coming from numerous similar distractor objects or individuals present in the test image. The third and fourth sub-tasks—question-answering and caption generation—require the model to recoginize the identities in the test images, then either respond to related questions or produce a comprehensive description. These represents the most difficult level of tasks, as the model should not only recognize each instance's identity in the image but also provide appropriate answers based on information related to their names, including states, actions, locations, and more. 
The samples of sub-tasks are shown in Figure \ref{fig:benchmark}.

\footnotetext[1]{The urls of animation images are shown in \href{https://github.com/jiyt17/IDA-VLM}{Github}.}




\subsection{Data Collection and Statistics}

In the construction of MM-ID benchmark, we initially collect images that meet the specified conditions. To evaluate the model's capability of recognizing instance-level information, the images used for testing need to contain multiple characters. Furthermore, each character requires distinctiveness, such as different actions, clothing attributes, etc., to facilitate the design of questions. Our data sources primarily encompass three types: shot images of movie scenes from the MovieNet dataset, animated images with download links collected from the internet, and pictures of buildings, vehicles, and animals gathered from object re-identification (Re-ID) datasets~\cite{DBLP:journals/corr/abs-2401-06960/reidobject, DBLP:conf/nips/JiaoLGWL0Z23/reidanimal}. In each example, there exists ID images corresponding to each character and one or more test images. 
 
Following preparing images of MM-ID, we annotate questions and standard answers manually. For every subtask, we use GPT-4 to generate templates, such as instructions for descriptions, localization. In question-answering tasks, MM-ID poses questions regarding specific characters' actions, attributes, clothing colors, positions, relationships, emotions, etc. After question annotation, we request other individuals' assistance in refinement and balancing, increasing the diversity of expression and grammatical structure. Finally, the correct answers are labeled manually by other annotators, providing accurate and detailed descriptions for caption task and correct answers for Q\&A task. 
Totally, we collect and annotate 585 samples. The data composition is shown in Appendix \ref{mm-id}.

\subsection{Evaluation Strategy}

Quantitative evaluation for open-domain LVLMs has always been challenging. Our MM-ID incorporates four sub-tasks, each presenting its unique answer format. 
For matching task, which resembles multiple-choice questions, accuracy is used as a metric. 
In localization task, we compare the predicted bounding box to the actual bounding box, computing the Intersection over Union (IoU) metric to measure accuracy. When IoU exceeds a threshold of 0.5, we consider the model recognizes instance identity accurately.

For question-answering and caption generation tasks, due to the open-ended nature of the generated responses, it is challenging to employ rule-based evaluations. Hence, GPT-4 is used to score the results. Under the condition of provided questions and correct answers, we design prompts to guide GPT-4 to focus on scoring the accuracy of character roles, states, and activities within predictions. We use both absolute and relative scoring strategies. Absolute scoring rates a model's prediction directly against the correct answer on a ten-point scale, while relative scoring pits two models' results against each other, providing more immediate comparative insights into the models' performance with a fraction. The prompts used for GPT-4 scoring are shown in Appendix \ref{mm-id}.

\section{Experiments}

\subsection{Quantitative Results}
We compare IDA-VLM with other open-source LVLMs~\cite{zhao2023mmicl, DBLP:journals/corr/abs-2310-01218/seed, DBLP:journals/corr/abs-2308-12966/qwen, DBLP:journals/corr/abs-2401-16420/intern} and closed-source APIs~\cite{DBLP:journals/corr/abs-2308-12966/qwen, gemini, gpt4v} on MM-ID. To meet the requirement of ID recognition, the models selected for testing should support multiple images input. For each model, we design appropriate prompts to guide them to respond according to character names. The prompts and training settings of our model can be found in Appendix \ref{settings}. 

As illustrated in Table \ref{tab: main result}, our model achieves the best performance on MM-ID. According to the quantitative results, Gemini-pro exhibits best performance on matching and Q\&A sub-tasks among previous models. In the location sub-task, QwenVL-Chat gains the highest score, even higher than other closed-source APIs. As for the caption sub-task, GPT-4V surpasses other models. Overall, our model outperforms previous LVLMs by a significant margin on four sub-tasks.

Because the absolute scoring of GPT-4 is not entirely objective, we calculate relative scores using GPT-4 to provide a further demonstration on the `Q\&A' and `Caption' sub-tasks. We compare the predictions from various closed-source APIs against those from our model respectively. All relative scores are below 1, signifying that our model outperforms its counterparts. Among the closed-source APIs evaluated, GPT-4V exhibits comparatively superior results.

\begin{table}[t]
\centering
\begin{adjustbox}{width=0.5\textwidth}

\begin{tabular}{ccccc}
\toprule
\multicolumn{1}{c|}{\textbf{Model}}      & \multicolumn{1}{l}{\textbf{Matching}} & \multicolumn{1}{l}{\textbf{Location}} & \multicolumn{1}{l}{\textbf{Q\&A}} & \multicolumn{1}{l}{\textbf{Caption}} \\ \midrule
\multicolumn{5}{l}{\textbf{Open-source Models}}                                                                                                                                                     \\ \midrule
\multicolumn{1}{c|}{MMICL}               & --                                    & --                                    & 3.53                              & 3.18                                 \\
\multicolumn{1}{c|}{SEED}                & --                                    & --                                    & 3.19                              & 3.58                                 \\
\multicolumn{1}{c|}{QwenVL-Chat}         & --                                    & 0.504                                 & 3.63                              & 2.65                                 \\
\multicolumn{1}{c|}{InternLM-XComposer2} &                  --                     &            0.106                &       3.44                            &                 2.93              \\ \midrule
\multicolumn{5}{l}{\textbf{Closed-source APIs}}                                                                                                                                                      \\ \midrule
\multicolumn{1}{c|}{QwenVL-Plus}         & 0.313                                 & 0.187                                 & 3.87                              & 3.79                                 \\
\multicolumn{1}{c|}{QwenVL-Max}          & 0.224                                 & 0.301                                 & 4.64                              & 4.23                                 \\
\multicolumn{1}{c|}{Gemini-pro}          & 0.687                                 & 0.081                                 & 4.97                              & 4.04                                 \\
\multicolumn{1}{c|}{GPT-4V}              & 0.627                                 & 0.244                                 & 4.77                              & 4.67                                 \\ \midrule
\multicolumn{1}{c|}{IDA-VLM}           & \textbf{0.716}                                 & \textbf{0.821}                                 & \textbf{5.71}                              & \textbf{5.15}              \\ \bottomrule
\end{tabular}
\end{adjustbox}
\caption{The comparison results on MM-ID. Notably, The scores of `Q\&A' and `Caption' are evaluated with GPT-4 absolute scoring strategy. `-' indicates the model can't complete corresponding instruction.}
\label{tab: main result}
\vspace{-0.2cm}
\end{table}

\begin{table}[]
\centering
\begin{adjustbox}{width=0.4\textwidth}
\begin{tabular}{c|cc}
\toprule
\textbf{Model}                          & \multicolumn{1}{l}{\textbf{Q\&A}} & \multicolumn{1}{l}{\textbf{Caption}} \\ \midrule
GPT-4V / IDA-VLM                           & 0.925                         & 0.793                            \\
\multicolumn{1}{l|}{Gemini-pro / IDA-VLM}  & 0.892                         & 0.704                            \\
\multicolumn{1}{l|}{QwenVL-Plus / IDA-VLM} & 0.796                         & 0.544                            \\
QwenVL-Max / IDA-VLM                       & 0.918                         & 0.636                            \\ \bottomrule
\end{tabular}
\label{tab: main compare}
\end{adjustbox}
\caption{Using relative scoring strategy to compare previous LVLMs with our model. Each result is the ratio of the first model's score to that of the second model's, after feeding the predictions of two models into GPT-4 for comparison.}
\vspace{-0.3cm}
\end{table}

\begin{figure*}[t]
	\centering
	\includegraphics[width=\textwidth]{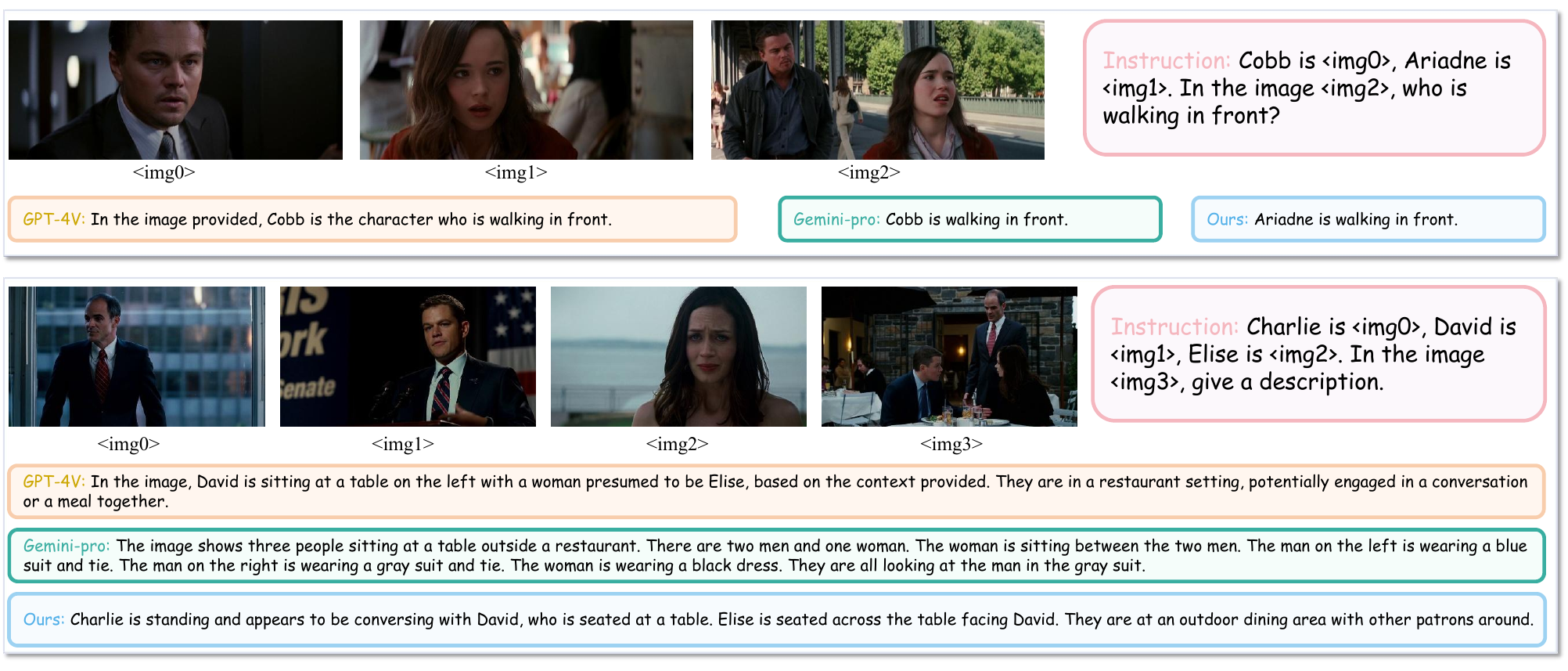}
	\caption{We present visualizations of selected samples from MM-ID, corresponding to Q\&A and caption sub-tasks. We showcase responses of GPT-4V, Gemini-pro and ours (IDA-VLM).}
	\label{fig:compare}
 \vspace{-0.2cm}
\end{figure*}

\subsection{Qualitative Comparison}

As shown in Figure \ref{fig:compare}, we qualitatively compare the results of IDA-VLM with GPT-4V and Gemini-pro. In the first case, the model should determine which person walks in front and match the person with the given characters. Our model gives the right answer `Ariadne' while GPT-4V and Gemini-pro recognizes inaccurately. In the second case, models are asked to provide a description for a movie scene with three characters. GPT-4V describes that David and Elise sit at a table, ignoring Charlie standing. Gemini-pro only gives a generic caption for the test image without recognizing characters' identities within the context. In contrast, our model gives the most reasonable description of the scene. 
We present more qualitative examples in Appendix \ref{visualization}.

\subsection{Ablation Studies}
\noindent\textbf{Effect of Dual-stage instruction tuning.}
We report the separate effect of each instruction tuning stage in Table \ref{tab:dual stage tuning}. When the second stage of tuning is removed, there is a significant drop in performance, suggesting that the second stage has a more substantial impact than the first. During the second stage, the ID images are independent of the test images, which enhances the quality of data for model learning. When the first and second stages of tuning are combined, our model achieves its optimal performance. It is worth noting that the accuracy of `Matching' gets lower when adding the first stage tuning, because the first stage tuning data can't improve matching ability.

\begin{table}[th]
\centering
\begin{adjustbox}{width=0.5\textwidth}
\begin{tabular}{c|cccc}
\toprule
\textbf{Model}          & \multicolumn{1}{l}{\textbf{Matching}} & \multicolumn{1}{l}{\textbf{Location}} & \multicolumn{1}{l}{\textbf{Q\&A}} & \multicolumn{1}{l}{\textbf{Caption}} \\ \midrule
w/o first stage tuning  &                \textbf{0.746}                 &                  0.813           &           5.43             &       5.08                           \\
w/o second stage tuning &    --                            & 0.715                        & 4.08        & 3.86    \\
Ours                    &          0.716                     &                \textbf{0.821}               &              \textbf{5.80}             &   \textbf{5.19}                              \\ \bottomrule
\end{tabular}
\end{adjustbox}
\caption{The ablation study about separate stage instruction tuning.\protect\footnotemark[1]}
\label{tab:dual stage tuning}
\vspace{-0.2cm}
\end{table}

\footnotetext[1]{Because the sample of the first stage tuning has only one text image, we present the `QA' and `caption' results on a sub-set of MM-ID, which has one test image in each sample.}

\begin{table}[th]
\centering
\begin{adjustbox}{width=0.5\textwidth}
\begin{tabular}{c|cccc}
\toprule
\textbf{Model} & \multicolumn{1}{l}{\textbf{Matching}} & \multicolumn{1}{l}{\textbf{Location}} & \multicolumn{1}{l}{\textbf{Q\&A}} & \multicolumn{1}{l}{\textbf{Caption}} \\ \midrule
w/o ID-Former  & 0.701                                 & 0.803                                 & 5.44                              & 5.13                                 \\
Ours           & \textbf{0.716}                                 & \textbf{0.821}                                 & \textbf{5.71}                              & \textbf{5.15}                                 \\ \bottomrule
\end{tabular}
\end{adjustbox}
\caption{The ablation study about ID-Former.}
\label{tab: id-former}
\vspace{-0.2cm}
\end{table}

\noindent\textbf{Effect of ID-Former.}
We use ID-Former to project visual features of ID images and test images to the semantic space of LLM. As depicted in Table \ref{tab: id-former}, substituting the ID-Former with a standard query former leads to a reduction in all sub-task metrics. The ID-Former enhances the interaction between features of ID images and test images, thereby activating ID information within the test images.

The ablation study for the mixing rate of LLaVA instruction tuning data are depicted in Appendix \ref{tuning data construction}.

\section{Conclusion}

In this paper, we focus on the capability to recognize and link instances across various scenes, which is significant for understanding complex visual narratives, such as movies. We propose visual instruction tuning with ID reference, which unleashes the potential of LVLM in ID recognition, and develop an ID-aware LVLM, IDA-VLM. 
The model memorizes the name and appearance of each character in ID reference, then recognize them across disparate scenes correctly, linking characters across diverse images for accurate narrative interpretation. 
To thoroughly assess the performance of instance recognition with ID reference, a novel benchmark named MM-ID has been introduced, which consists of four sub-tasks. Our model achieves best performance among previous models and closed-source APIs. 
Conclusively, this research contributes to broadening the horizons for future AI systems to efficiently understand multi-identity visual content. 

\section{Limitations}

As we employ QwenVL-chat as the baseline model for our research, our proposed model inherits certain limitations thereof. One such constraint is the upper limit on the number of input images. Within a sample, the sum count of ID images and test images can not surpass eight. Consequently, we opt to use keyframes as test images to effectively perform comprehension of movie segments. By integrating our visual instruction tuning with ID reference, if we fortify the baseline, which is anticipated to enhance overall performance. 

Our model focuses on the capability to recognize and link instance IDs across various scenes for accurate narrative interpretation. It should be acknowledged that during the visual instruction tuning with ID reference, there might be a slight reduction in the original capabilities of the model. However, our objective is to develop an ID-aware LVLM, rather than a versatile LVLM. 

\section{Ethical Considerations}

Our method is related to person and instance ID, so we pay much attention when collecting data. 
Our tuning data and benchmark data mainly come from MovieNet, which is a open-source movie dataset, containing numerous shot images of actors, necessitating careful consideration about the copyright. 
Moreover, there exists the potential risk that our model or dataset may be utilized by others to engage in activities concerning a particular individual.
\bibliography{custom}
\clearpage
\appendix

\section{Visual Instruction Tuning Data Construction}
\label{tuning data construction}

We construct dual-stage instruction tuning data with ID reference. The first stage data is constructed with predefined rules, cropping specific instances from test images, while the second stage data is generated by GPT-4V. Specifically, GPT-4V is utilized to produce four kinds of tuning datasets: single-image caption, multi-image caption, question-answer pair for a single image, and question-answer pair for multiple images. We present prompts for producing data in Table \ref{tab:gpt4v anno1} and Table \ref{tab:gpt4v anno2}.
The constructed tuning data scale of two stages are shown in Table \ref{tab: id rec data}. We will produce more data and open-source them.

\begin{table*}[]\centering
\begin{minipage}{1.0\textwidth}\vspace{0mm}    \centering
\begin{tcolorbox} 
    \centering
    \begin{tabular}{p{0.99\textwidth} c}
   \VarSty{ {\bf Description generation for a single test image} } &\\
You are a helpful and precise assistant for providing a description for an appropriate image. \\
User will give an image and the image size (width, height). Then user will give some character names and their position in the image. The position is expressed with bounding box, which is the person's left-top corner coordinates and right-bottom corner coordinates (left, top, right, bottom). \\
Firstly, you need to judge if it is appropriate. An appropriate image should be clear, should be easy for you to give a caption, the people in it should be easy to recognize. \\
If the image is not appropriate, you should answer `no', if it is appropriate, you should give an accurate description of the image with given character names according to the following rules. \\
Different characters will be split by `\textbackslash n', you must remember the right people in the right position. Maybe there are some other people without name in the image, your caption need to contain them if necessary, but the main subject should be about characters with names. \\
The description should be accurate and brief. The answer should be less than 60 words. Please pay more attention to people's action. Your answer should be only about the visual content, don't include your own speculation. \\
Then you should give an accurate description of the image with given character names.
& \\
& \\
\VarSty{ {\bf Description generation for multiple test images } } & \\
You are a helpful and precise assistant for providing a description for some continuous images. You can treat it as video caption generation. You need give an overall story description for these images according to the following rules. \\
User will give the image size (width, height) and some images. For each image, user will give some character names and their positions in the image. The position is expressed with bounding box, which is the person left-top corner coordinates and right-bottom corner coordinates (left, top, right, bottom). \\
Different characters will be split by '\textbackslash n', you must remember the right people in the right position. Maybe there are some other people without name in the image, your caption need contain them if necessary, but the main subject should be about characters with names. \\
The description should be accurate and brief. The answer should be less than 60 words. Please pay more attention to people's action. Your answer should be in accordance with temporal order of the images. Your description should be coherent and have some logical connections. Your answer should be only about the visual content, don't include your own speculation. \\
You should describe the changes in the characters' states and interactions throughout the entire images sequence as much as possible, avoiding fragmented descriptions for each individual image. Especially refrain from using phrases like 'in the image 1' and so on. \\
Then you should give an accurate and brief description of the images with given character names. \\
A good example: 'Carol watches as Hal, in a white tank top, looks at his shirt. As she approaches, they engage in a close conversation, and eventually, Hal looks at Carol while gesturing, continuing their discussion.' & \\


    \end{tabular}
\end{tcolorbox}
\caption{Prompts for GPT-4v to annotate `Caption' instruction tuning data based on MovieNet.}
    \label{tab:gpt4v anno1}
\end{minipage}
\end{table*}

\begin{table*}[]\centering
\begin{minipage}{1.0\textwidth}\vspace{0mm}    \centering
\begin{tcolorbox} 
    \centering
    \begin{tabular}{p{0.99\textwidth} c}
\VarSty{ {\bf Q\&A generation for a single test image} } & \\
You are a helpful and precise assistant for providing a question-answer pair of an image with given character names. \\
User will give an image and the image size (width, height). Then user will give some character names and their position in the image. The position is expressed with bounding box, which is the person left-top corner coordinates and right-bottom corner coordinates (left, top, right, bottom). \\
Firstly, you need to judge if it is appropriate. An appropriate image should be clear, should be easy for you to give a caption, the people in it should be easy to recognize. \\
If the image is not appropriate, you should answer 'no', if it is appropriate, you should give a question-answer pair of the image with given character names according to following rules. \\
Different characters will be split by '\textbackslash n', you must remember the right people in the right position. \\
Then you should give a pair of question and corresponding answer about the image with given character names. The question and answer should be split by '\textbackslash n'. \\
The question asks about the given character, including character actions, character attributes (clothes, expression, etc), character locations, relative relationship between characters, etc. Only include questions that have definite answers. Some examples of question templates: What is xxx doing? What color is xxx's clothes? \\
The question and answer should be accurate and brief. The answer should be strictly correspond to the question and be less than 30 words.   & \\
\\
\VarSty{ {\bf Q\&A generation for multiple test images}} & \\
You are a helpful and precise assistant for providing a question-answer pair for some continuous images with given character names. You can treat it as video queation answering generation. You need give an overall question and answer for these images according to the following rules. \\
User will give the image size (width, height) and some images. For each image, user will give some character names and their positions in the image. The position is expressed with bounding box, which is the person left-top corner coordinates and right-bottom corner coordinates (left, top, right, bottom). \\
Different characters will be split by '\textbackslash n', you must remember the right people in the right position. \\
Then you should give a pair of question and corresponding answer about the images with given character names. The question and answer should be split by '\textbackslash n'. \\
The question asks about one of or some given characters, including character actions, character attributes (clothes, expression, etc), relative relationship between characters, etc. Only include questions that have definite answers. \\
The question and answer should be accurate and brief. The answer should be strictly correspond to the question and be less than 30 words. \\
You should focus on the changes in the characters' states, actions or interactions throughout the entire images sequence as much as possible, avoiding fragmented question for each individual image. Especially refrain from using phrases like 'in the image 1' and so on. \\
A good example: 'What is Timmy doing?\textbackslash nTimmy walks into the room, then has a conversation with another man, finally they hug each other excitedly.' & \\

    \end{tabular}
\end{tcolorbox}
\end{minipage}
\caption{Prompts for GPT-4v to annotate `Q\&A' instruction tuning data based on MovieNet.}
    \label{tab:gpt4v anno2}
\end{table*}

To preserve the innate proficiency of the baseline model, we integrate the instruction tuning dataset from LLaVA with our compiled dataset. As depicted in Table \ref{tab:ablation llava}, we conduct an ablation study on the proportion of LLaVA dataset. The capability for `Matching' reaches its best in the absence of LLaVA dataset, suggesting LLaVA data has no benefit for `Matching'. The location score is highest at a 20\% inclusion rate of LLaVA data. In contrast, `Q\&A' and `Caption' scores reach their optimal levels when the inclusion rate is set at 10\%. This could imply that a judicious amount of LLaVA data serves to fortify foundational multimodal and text generation competencies. However, an excessive infusion of LLaVA data seems to hinder the learning of ID recognition. Therefore, we select 10\% as the mixing rate of LLaVA data.

\begin{table*}[t]
\centering
\begin{tabular}{ccc}
\toprule
\textbf{Stage}                             & \textbf{tuning data} & \textbf{data scale} \\ \midrule
\multirow{3}{*}{\textbf{The first stage}}  & VCR                  & 60k                  \\
                                           & RefCoco              & 60k                  \\
                                           & Flickr30k            & 60k                 \\ \midrule
\multirow{6}{*}{\textbf{The second stage}} & Matching             & 10k                  \\
                                           & Location             & 10k                  \\
                                           & Single-image Q\&A                   & 10k                \\
                                           & Single-image Caption              & 10k                \\
                                           & Multi-image Q\&A       & 10k                  \\
                                           & Multi-image Caption  & 10k                  \\ \bottomrule
\end{tabular}
\caption{ID recognition instruct-tuning dataset.}
\label{tab: id rec data}
\end{table*}

\begin{table*}[th]
\centering
\begin{tabular}{c|cccc}
\toprule
\textbf{Model}         & \multicolumn{1}{l}{\textbf{Matching}} & \multicolumn{1}{l}{\textbf{Location}} & \multicolumn{1}{l}{\textbf{Q\&A}}   & \multicolumn{1}{l}{\textbf{Caption}} \\ \midrule
Ours (w/o LLaVA data)  &  \textbf{0.746} &  0.797   &   5.27    &      4.98        \\
Ours (10\% LLaVA data) &  0.716                                 & 0.821                                 & \textbf{5.71}                              & \textbf{5.15}              \\
Ours (20\% LLaVA data) &  0.716 &   \textbf{0.829}   &  5.67   &      5.11               \\ \bottomrule
\end{tabular}
\caption{The ablation study about instruction tuning data of LLaVA.}
\label{tab:ablation llava}
\end{table*}

\section{MM-ID}
\label{mm-id}

The data composition of MM-ID is shown in Table \ref{tab: mmid statistic}. 
During MM-ID construction, we annotate questions and standard answers manually. The instructions for annotators are shown in Table \ref{tab:annotate instruct}.

It is a challenge to evaluate accuracy of the model on `Q\&A' and `Caption' sub-tasks. We utilize GPT-4 to score the predictions under the condition of provided questions and correct answers. We propose GPT-4 absolute and relative scoring strategies with designed prompts, as depicted in Table \ref{tab:gpt4 eval}.

The calculate process of relative score is as follows: GPT-4 gives two scores (10 points) for predictions from different models respectively. We compute mean value of scores for each model, then calculate the ratio of the average scores as the final relative score.

\begin{table*}[th]
\centering
\begin{tabular}{cccc}
\toprule
\textbf{Sub-task} & \multicolumn{1}{l}{\textbf{Question Num}} & \textbf{Image Num} & \textbf{Source} \\ \midrule
Matching          & 67                   &    335   &    MovieNet, Re-id datasets    \\
Location          & 123                  &    360   &    MovieNet, animations    \\
Q\&A              & 154                  &    507   &    MovieNet, animations    \\
Caption           & 241                  &    1110   &    MovieNet, animations    \\ \bottomrule
\end{tabular}
\caption{MM-ID data statistic.}
\label{tab: mmid statistic}
\end{table*}

\begin{table*}[t]\centering
\begin{minipage}{1.0\textwidth}\vspace{0mm}    \centering
\begin{tcolorbox} 
    \centering
    \begin{tabular}{p{0.97\textwidth} c}
Design Questions: The questions are about specific characters in an image. These questions require the model to correctly identify a specific person to provide an answer, such as inquiring about someone's condition, actions, positions, etc. Consider more angles for asking questions.

Annotate Answers for Q\&A and Descriptive Types of Questions: Answers to Q\&A types of questions need to be accurate based on the character. Descriptive questions only require a correct description, as detailed as possible. Besides what each person is doing, include some simple background descriptions as well.

    \end{tabular}
\end{tcolorbox}
\end{minipage}
\caption{The instructions for Annotators.}
    \label{tab:annotate instruct}
\end{table*}

\begin{table*}[t]\centering
\begin{minipage}{1.0\textwidth}\vspace{0mm}    \centering
\begin{tcolorbox} 
    \centering
    \begin{tabular}{p{0.97\textwidth} c}

\VarSty{ {\bf Absolute Score} } & \\
You are a helpful and precise assistant for evaluating answers. We would like to request your feedback on the quality of an AI assistant's answer according to the given question and ground truth. The question, answer of AI assistant and ground truth will be signed by `question', `prediction' and `GT'. You need to judge whether the overall meanings of prediction and ground truth answer are consistent or not. Please pay more attention to the correspondence of character names and their states or actions. Please rate the helpfulness, relevance, accuracy of the responses. You should give an overall score on a scale of 1 to 10, where a higher score indicates better overall performance. Please first output a single line containing only one value indicating the score for Assistant answer. In the subsequent line, please provide a comprehensive explanation of your evaluation, avoiding any potential bias and ensuring that the order in which the responses were presented does not affect your judgment. & \\
\\
\VarSty{ {\bf Relative Score} } & \\
We would like to request your feedback on the performance of two AI assistants in response to the user question according to the given ground truth. The question, ground truth answer and predictions of two AI assistants will be signed by 'question', 'GT', 'prediction 1' and 'prediction 2'. Please rate the helpfulness, relevance, accuracy, level of details of their responses. Each assistant receives an overall score on a scale of 1 to 10, where a higher score indicates better overall performance. Please pay more attention to the correspondence of character names and their states or actions. Please first output a single line containing only two values indicating the scores for Prediction 1 and 2, respectively. The two scores are separated by a space. In the subsequent line, please provide a comprehensive explanation of your evaluation, avoiding any potential bias and ensuring that the order in which the responses were presented does not affect your judgment. & \\

    \end{tabular}
\end{tcolorbox}
    
\end{minipage}
\caption{The prompts for evaluating with GPT-4 on MM-ID.}
\label{tab:gpt4 eval}
\end{table*}

\section{Experiment Settings}
\label{settings}

We use QwenVL-chat for model initialization, which has 9.6B parameters. In IDA-VLM training, we set learning rate as 1e-5 for the first stage and 5e-6 for the second stage. The model is trained for 5 epochs in both the first and second stages. The batch size with gradient accumulation is set to 128. The visual encoder is fixed, while ID-Former and LLM are fine-tuned. It costs approximately 1 day training on 8$\times$A100-SXM-80GB for both the first stage and second stage tuning.

To promote the evaluation of models on MM-ID tasks, we craft specific prompts to instruct models in accurately identifying character IDs. The prompts tailored for closed-source APIs are detailed in Table \ref{tab:api test}. and prompts for open-source models are designed similarly.

\begin{table*}[t]\centering
\begin{minipage}{1.0\textwidth}\vspace{0mm}    \centering
\begin{tcolorbox} 
    \centering
    \begin{tabular}{p{0.97\textwidth} c}
\VarSty{ {\bf Matching, Q\&A, Caption (GPT-4V, Gemini-pro, QwenVL-Plus, QwenVL-Max)} } & \\
You are a helpful and precise assistant for providing a answer to the question. You need recognize instance identity to answer questions about reference characters or give a caption with character names (for multiple continuous images). You must provide an exact answer. & \\
\\
\VarSty{ {\bf Location (GPT-4V, Gemini-pro)} } & \\
You need to give coordinates of bounding box of some given characters or objects. The answer form should be `bbox: [x1, y1, x2, y2].', where x1 is left side of bounding box, y1 is upper side, x2 is right side, y2 is bottom side, they are all integers. & \\
\\
\VarSty{{\bf Location (QwenVL-Plus, QwenVL-Max)}} & \\
You need to give coordinates of bounding box of one given character or object. The answer form should only be '<ref>xxx</ref><box>(x1,y1),(x2,y2)</box>.', where x1 is left side of bounding box, y1 is upper side, x2 is right side, y2 is bottom side, they are all integers. & \\
    \end{tabular}
\end{tcolorbox}
\end{minipage}
\caption{The prompts for closed-source APIs testing on MM-ID.}
    \label{tab:api test}
\end{table*}

\section{More Visualization}
\label{visualization}

\begin{figure*}[t]
	\centering
	\includegraphics[width=\textwidth]{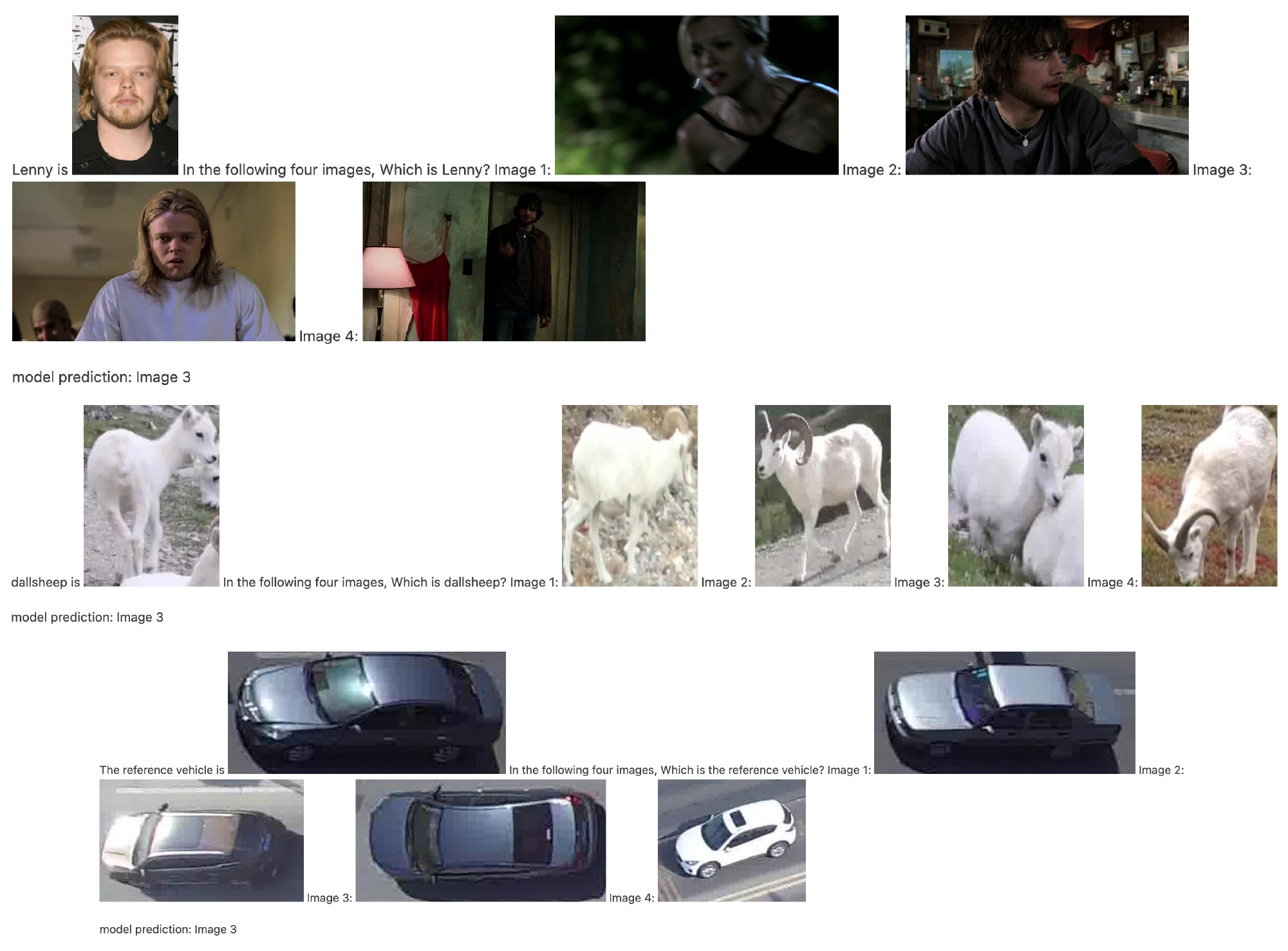}
	\caption{Samples of Matching sub-task.}
	\label{fig:appendix1}
\end{figure*}

\begin{figure*}[t]
	\centering
	\includegraphics[width=\textwidth]{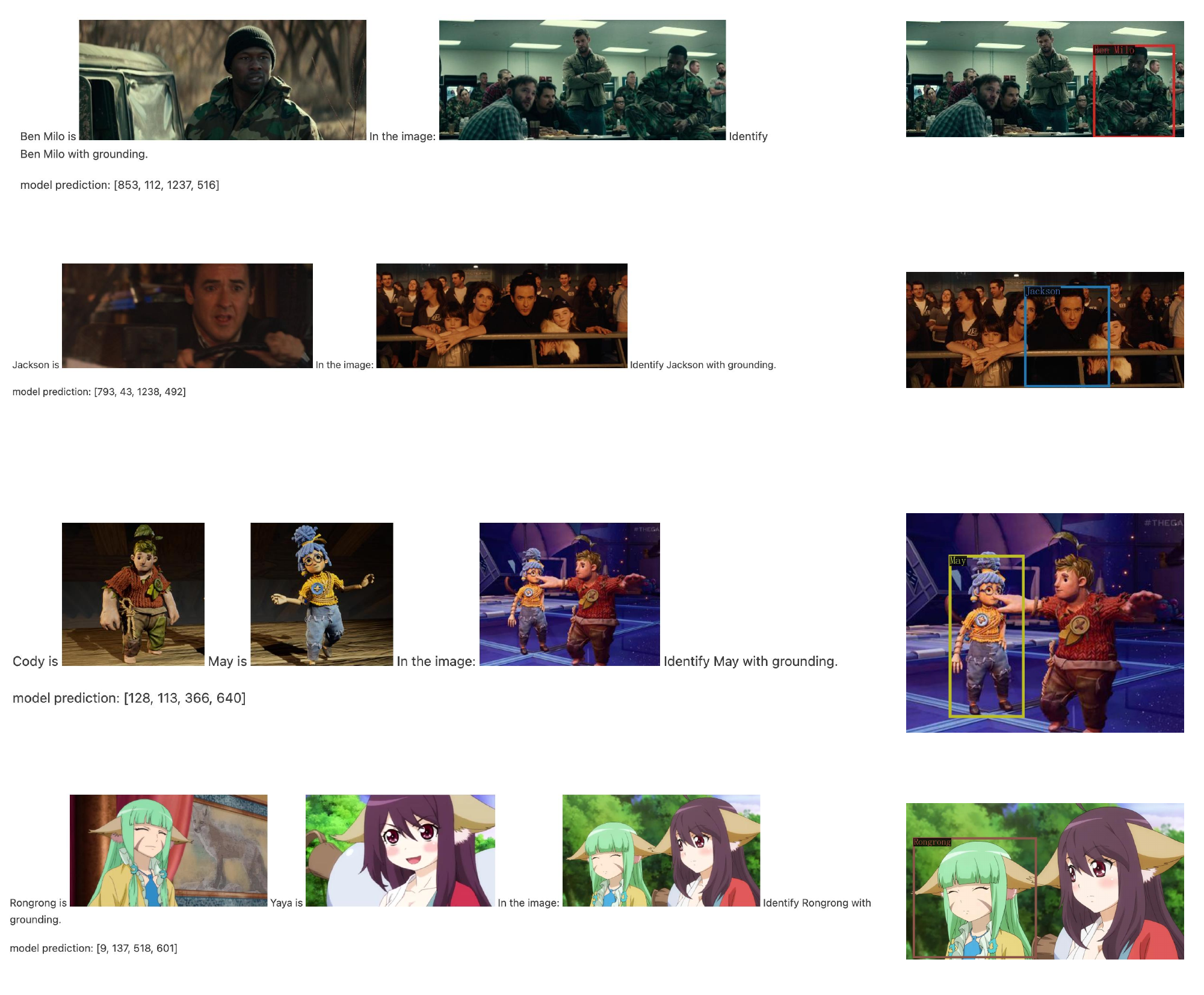}
	\caption{Samples of Location sub-task.}
	\label{fig:appendix2}
\end{figure*}

\begin{figure*}[t]
	\centering
	\includegraphics[width=\textwidth]{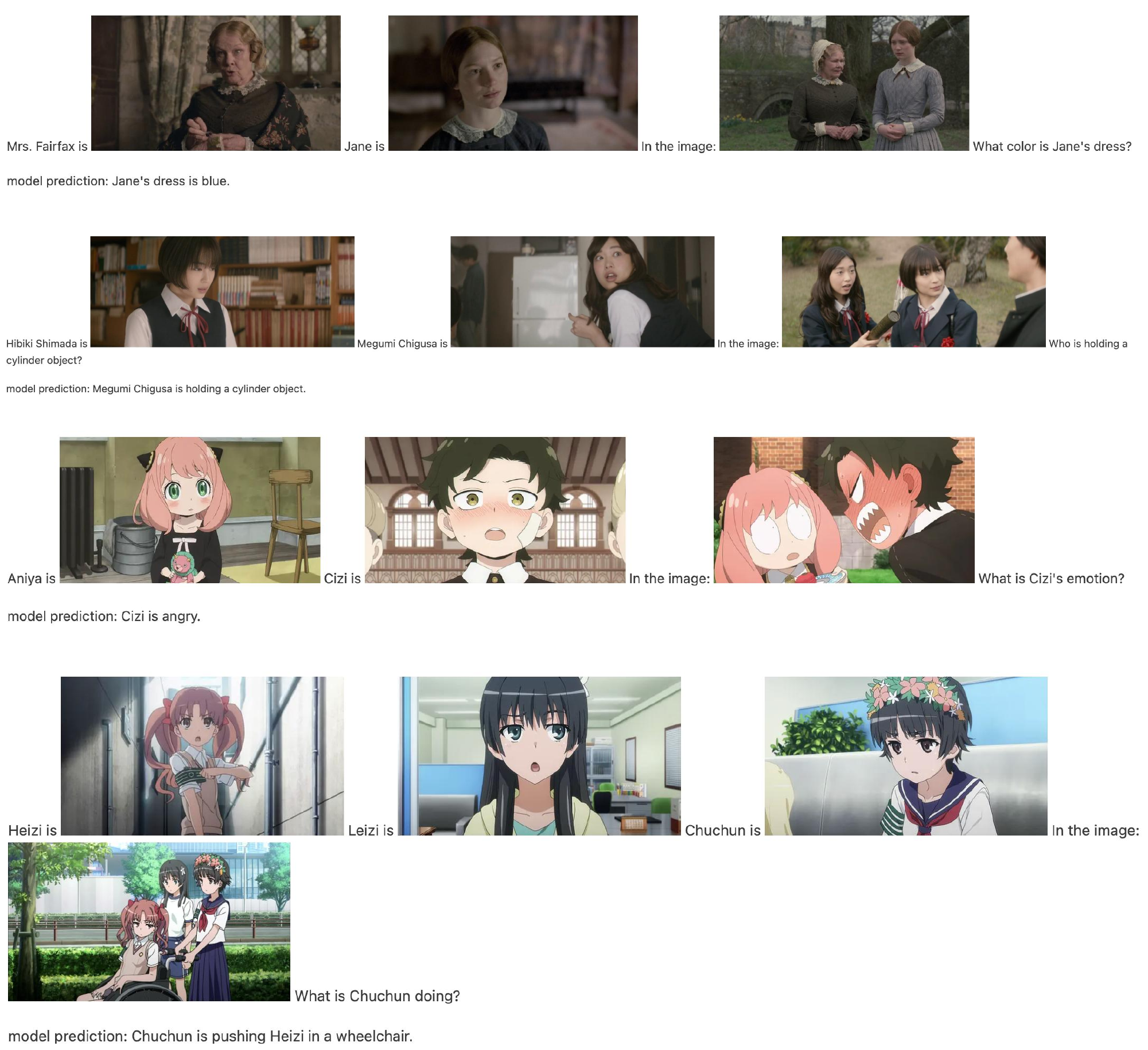}
	\caption{Samples of Q\&A sub-task.}
	\label{fig:appendix3}
\end{figure*}

\begin{figure*}[t]
	\centering
	\includegraphics[width=\textwidth]{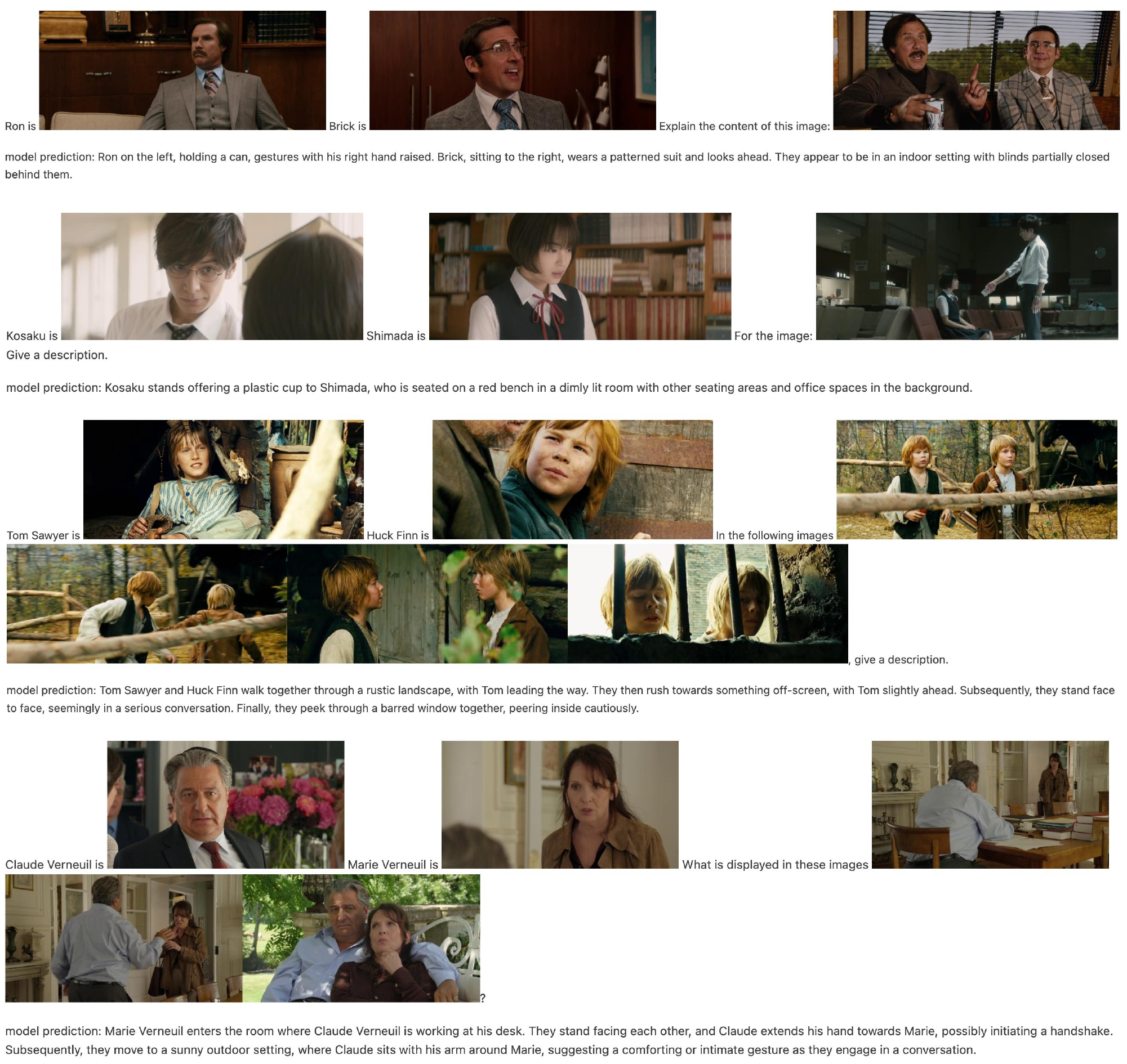}
	\caption{Samples of Caption sub-task.}
	\label{fig:appendix4}
\end{figure*}

In this section, we present more qualitative results of IDA-VLM, as shown in Figure \ref{fig:appendix1}, Figure \ref{fig:appendix2}, Figure \ref{fig:appendix3}, Figure \ref{fig:appendix4}. 

\end{document}